\documentclass{Foresj}

\usepackage{etoolbox}
\makeatletter
\patchcmd{\@maketitle}{\vspace*{24pt}}{\vspace*{-24pt}}{}{}
\makeatother

\usepackage{etoolbox}
\makeatletter

\let\titlepageheadline\relax

\renewcommand\titlepagefootline{\rlap{\rule[21\p@]{\textwidth}{0.5\p@}}\hfill\thepage}

\def\@journal@title{\protect\leavevmode\protect\vphantom{Forestry}}

\def\@oddfoot{\rlap{\rule[21\p@]{\textwidth}{0.5\p@}}\hfill\thepage}
\def\@evenfoot{\rlap{\rule[21\p@]{\textwidth}{0.5\p@}}\thepage\hfill}

\makeatother

\usepackage{booktabs}

\begin{document}
\nolinenumbers
\doublespace

\title[Sparse tensor networks for tree diameter estimation]{Sparse tensor networks for individual tree diameter estimation in mixed forests}

\author[Lößl, Fassnacht and Barkov]{%
  Clara \surname{L\"{o}\ss l}$^{1,2,3,\ast}$,
  Fabian Ewald \surname{Fassnacht}$^{4}$ and
Viacheslav \surname{Barkov}$^{1,2}$}

\address{%
  $^{1}$Joint Lab Artificial Intelligence
  and Data Science, Osnabr\"{u}ck University, Osnabr\"{u}ck, Germany\\
  $^{2}$Department of Agromechatronics, Leibniz Institute for Agricultural Engineering and Bioeconomy
  (ATB), Potsdam, Germany\\
  $^{3}$Chair of Agromechatronics, Technische Universität Berlin, Berlin, Germany\\
  $^{4}$Institute of Geomatics, University of Natural Resources and Life Sciences (BOKU), Vienna, Austria}

\corres{$^{\ast}$Corresponding author. E-mail: clara.loessl@uni-osnabrueck.de}

\begin{abstract}
  Measurements of tree diameter at breast height (DBH) are fundamental to forest inventories, but manual surveys across entire stands or regions are impractical. UAV-borne laser scanning (ULS) enables scalable automated forest surveys, yet canopy occlusion makes accurate DBH estimation from ULS point clouds non-trivial. Limited stem-level point density of ULS data constrains approaches relying on detailed stem reconstruction, such as Quantitative Structure Models (QSMs) and circle fitting. Alternatively, empirical machine learning methods based on predefined point cloud metrics are applied. New deep learning approaches, however, offer a more principled alternative, allowing models to learn directly from point clouds. Sparse tensor networks have shown particular promise, already transforming forestry tasks such as tree species classification. We investigate whether these strengths can be extended to estimate individual tree DBH from ULS data across four mixed-species temperate forest plots in Germany. We further advance existing sparse tensor networks by introducing TreeNext, a lightweight deep learning architecture for direct individual tree parameter estimation from voxelized LiDAR point clouds. We present a fully automated forest inventory pipeline from individual tree instance segmentation to DBH regression, and quantify the impact of segmentation errors by comparing performance on automatically segmented and ground truth instances. TreeNext achieves an $R^2$ of 0.8355 on automatically segmented trees and 0.9380 on ground truth segments, outperforming strong tabular baselines including Random Forest and XGBoost. Segmentation quality emerges as the dominant error source in the automated pipeline. TreeNext proves highly data-efficient, performing well even in small data scenarios. Post-hoc ensembling yields further gains, allowing TreeNext to surpass the best-performing baseline using only 40\% of the available training data. Our findings establish sparse tensor networks as a practical, accurate foundation for automated individual tree DBH estimation from ULS, and a promising alternative when detailed woody-structure reconstruction is infeasible. Data and code are openly available.
\end{abstract}

\maketitle

\section{Introduction}
\label{section:introduction}

Forests are essential for climate regulation, biodiversity, and human well-being, providing a large and persistent global carbon sink and a wide range of ecosystem services \citep{watson2018exceptional,pan2011large}. Accurate measurement and estimation of individual tree parameters underpins modern forest inventories, enabling operational assessment of stand structure, management, and biodiversity indicators \citep{oettel2021linking}. Among these parameters, diameter at breast height (DBH) is fundamental. DBH is the standard, non-destructive, and repeatable measure of tree size that supports efficient data collection and long-term monitoring across diverse forest types \citep{liang2022close}. Critically, DBH is the strongest single predictor in allometric models and often suffices by itself to estimate many key forest attributes with high accuracy, most notably above-ground biomass \citep{brede2022nondestructive,kuyah2012allometric}.

Traditionally, DBH is measured manually using tools like diameter tapes, calipers, or dendrometers \citep{liang2022close}. While accurate, these field-based methods are widely understood as labor-intensive, time-consuming, and costly, limiting their application for large-scale inventories \citep{liang2022close, sun2022predicting, du2023lidar}. Remote sensing technologies offer efficient alternatives for estimating forest attributes \citep{ozturk2026barriers, chave2019ground}.

In particular, Light Detection and Ranging (LiDAR) technology enables the acquisition of high-density three-dimensional (3D) point clouds, allowing for the detailed characterization of complex forest structures due to its ability to penetrate the canopy \citep{brede2017comparing, sun2022predicting, du2023lidar}. This canopy penetration provides key advantages over photogrammetric methods (e.g., Structure from Motion), enabling more accurate tree measurements \citep{hartley2025uav}. Various laser scanning platforms have been developed to address different operational requirements and trade-offs between coverage, resolution, and deployment effort. Ground-based platforms, such as terrestrial laser scanning (TLS) and mobile laser scanning (MLS), provide highly detailed point clouds \citep{holvoet2025terrestrial, bekkers2014improving, brede2017comparing}. However, both TLS and MLS are labor-intensive for large plots, requiring multiple scan positions \citep{brede2022nondestructive, wilkes2017data} or narrow trajectories \citep{kuzelka2025trajectory} to mitigate occlusion, and face limitations in efficiently covering large areas and capturing the upper tree canopy \citep{dhont2024evaluating, liang2022close, brede2017comparing}. Airborne laser scanning (ALS) systems efficiently cover larger areas \citep{vezeteu2025direct}, with UAV-borne laser scanning (ULS) emerging as a particularly effective modern approach for smaller continuous forest areas \citep{brede2017comparing, du2023lidar}. ULS combines flexible deployment with the ability to acquire point clouds with point density substantially exceeding that of conventional manned ALS \citep{dersch2023towards,brede2022nondestructive,brede2017comparing}, enabling detailed tree analysis even in complex, mixed forests \citep{sun2022predicting, brede2022nondestructive}.

The capacity of ULS for rapid acquisition of high-quality 3D data opens a direct path to characterizing forests at large spatial scales with substantially reduced field effort. The resulting point clouds represent forest structure in high detail, and translating them into quantitative inventory measurements is a critical step toward highly automated forest inventories at single tree level. Such measurements underpin complex silvicultural decision-making, from selective thinning, fertilization, and herbicide application to the modeling of biomass and carbon storage at scales relevant for climate reporting \citep{oettel2021linking}, and emerging carbon-crediting markets, where forest owners may benefit from documenting actual biomass stocks at high spatial detail \citep{wheeler2026path}.

Multiple approaches have been developed to translate point cloud data into operational forest inventory products. Area-based approaches, among the earliest to reach operational maturity, rely on statistical models that predict average attributes for entire plots \citep{coops2021modelling, xu2021lidar}. While well suited for large-area monitoring and strategic planning \citep{coops2021modelling}, these methods inherently aggregate data, obscuring the fine-scale structural details of individual trees. The richness of modern LiDAR-derived point clouds, however, provides the opportunity to move beyond plot-level summaries to the scale of individual trees. Consequently, recent research has increasingly focused on individual tree approaches \citep{shao2024largescale}, which more closely resemble classical forest inventory and can yield detailed structural and demographic information, enabling practitioners to quantify competitive interactions within stands, assess growth trajectories, and inform tree-level management prescriptions \citep{xu2021lidar, rodrigues2024applications}. Achieving this level of detail from point cloud data is, however, a substantially more complex task, for which machine learning methods have become a common methodological solution.

Automating single-tree inventory with machine learning typically consists of two stages: individual tree segmentation and per-tree parameter estimation. The latter is commonly framed as a supervised machine learning task on structured, tabular data derived from the point cloud. For extremely high-density ground-based acquisitions such as TLS, this structured data can contain direct geometric proxies of the target parameter estimated from stem geometry, for example by circle fitting \citep{liang2022close}, or through quantitative structure models (QSMs) and related approaches that reconstruct woody tree structure \citep{demol2022estimating}. ULS, while offering the key advantages of rapid and flexible large-area coverage, does not generally permit such direct measurements due to canopy occlusion and lower stem-level point density \citep{brede2017comparing}. Instead, the principled approach for ULS-based tree parameter estimation relies on computing comprehensive sets of per-tree point cloud metrics (e.g., height percentiles, distribution statistics, return ratios) that describe overall tree structure rather than being limited to the geometry of a parameter of interest alone. These metrics then serve as input to common machine learning regressors such as Linear Regression, Random Forest, and gradient boosted decision trees (GBDT) \citep{ye2025estimating,xu2025improving,hartley2025uav,zhang2022estimation,sun2022predicting,du2023lidar}. While effective and widely adopted, incorporating only features that practitioners explicitly define can overlook further valuable 3D spatial context present in the raw point cloud, and can be especially sensitive to segmentation artifacts and varying data acquisition conditions \citep{rodrigues2024applications}. These limitations raise the question whether a model capable of learning directly from the raw 3D point cloud structure, bypassing explicit metric definition, could extract richer geometric and structural information and ultimately yield a more principled, resilient, and accurate approach to automated forest inventory. To realize this potential in the complex operational environments characteristic of forest ecosystems, modern deep learning represents a highly promising direction.

Deep learning has transformed classification and segmentation of forest point clouds \citep[e.g.][]{puliti2025benchmarking,murray2025individual,wielgosz2024segmentanytree}, yet direct regression of tree parameters from LiDAR-derived 3D point cloud data remains underexplored. Recent work explored dense 3D convolutional neural networks for plot-level above-ground biomass estimation from ALS data, achieving performance comparable to tabular baselines such as Random Forest but without consistently outperforming them \citep{schaefer2024cnn}. Advancements in data representation, particularly sparse tensor networks that exploit the inherent sparsity of point cloud data \citep{choy2019minkowskiengine,graham2018submanifold}, have since led to stronger performance, as demonstrated by \citet{oehmcke2024deep} for estimating plot-level biomass from ALS. However, whether these advantages extend to estimating individual tree parameters from ULS point clouds remains an open question, as this setting differs fundamentally in spatial scale, point density, and structural complexity. To date, sparse tensor networks have not been applied to single-tree DBH estimation, nor has deep learning been systematically evaluated for per-tree parameter regression from ULS point clouds, a need highlighted as one of the logical next steps toward improving single-tree inventory in precision forestry \citep{rodrigues2024applications}.

Here, we aim to address this need in several ways. First, we investigate whether sparse tensor networks, which have demonstrated strong performance for plot-level regression from ALS, can transfer to the complex domain of ULS data and the fine spatial scale of individual trees. Building upon these initial investigations, we aim to push the performance boundary further by introducing TreeNext, a lightweight sparse tensor network for forestry applications, incorporating improvements in model architecture, data representation, and computational efficiency. We ground our investigation in the operational realities of modern forest inventory by evaluating a fully automated deep learning pipeline, from individual tree segmentation to DBH regression. We further investigate the performance of the proposed method on both automatically segmented trees and ground truth tree segments to quantify segmentation-induced error propagation. Additionally, we examine the data efficiency of the proposed approach across varying training set sizes, addressing the common concern that acquiring high-quality ground truth data in forestry applications is expensive and labor-intensive. We present our results with fully open data and code to ensure reproducibility and facilitate future operational research.

\section{Materials and methods}
\label{section:materials-and-methods}

\subsection{Study area}
\label{section:study-area}

\begin{figure}[htbp]
  \centering
  \includegraphics{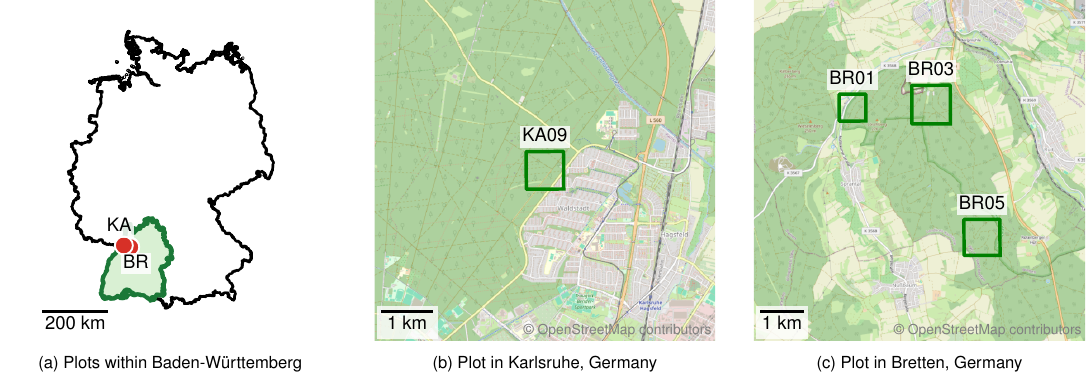}
  \caption{Geographic location of the study plots. (a) Location of Karlsruhe (KA) and Bretten (BR) within Baden-Württemberg, Germany. (b) Map of the Karlsruhe plot (KA09). (c) Maps of the three Bretten plots (BR01, BR03, BR05). The plots are 1-ha forest stands used throughout the study. Base map copyright: OpenStreetMap contributors.}
  \label{fig:study-areas}
\end{figure}

Our data represents typical Central European mixed temperate forests collected in Germany, openly available as part of the laser scanning dataset SYSSIFOSS \citep{weiser2022syssifoss-essd,weiser2022syssifoss-pangaea}. SYSSIFOSS addresses the need for standardized open access datasets in point cloud-based forestry research \citep{hartley2024tree} and has established itself as a commonly used benchmark in recent literature, serving as a standard for tasks ranging from individual tree segmentation and species classification to 3D forest scene reconstruction \citep[e.g.][]{esmoris2024deep,li2025individual,zhang2025hmemd,vahrenhold2025mmtscnet,xu2024threedimensional}.

The study sites are located in Baden-Württemberg, southwest Germany, within the municipal forests of Bretten (Kraichgau hills) and Karlsruhe Hardtwald (Upper Rhine Plain). Forest stands are mixed and multi-layered, dominated by Scots pine (Pinus sylvestris), European beech (Fagus sylvatica), oaks (Quercus spp.), Douglas fir (Pseudotsuga menziesii), and Norway spruce (Picea abies). The study sites comprise four 1-ha plots, with data acquired by ULS under leaf-off conditions \citep{weiser2022syssifoss-essd,weiser2022syssifoss-pangaea}. These plots are referenced by the names BR01, BR03, BR05 for Bretten plots and KA09 for the Karlsruhe plot throughout this manuscript. The locations of the plots are shown in Figure~\ref{fig:study-areas}.

\subsection{Data collection}
\label{section:data-collection}

The study relies on ULS point clouds and ground truth DBH measurements for validation. ULS acquisitions were conducted with a RIEGL miniVUX-1UAV mounted on a DJI Matrice 600 Pro, using two overlapping grid patterns with the second rotated by $45^\circ$ to increase multi-angle sampling. General ULS settings comprised pulse repetition frequency 100~kHz, scan angle $\pm90^\circ$, typical altitude ~60--75~m~a.g.l., speed 3--5~m s$^{-1}$, and flight line spacing 25--30~m. Trajectories were post-processed with Applanix POSPac MMS and point clouds were processed in RIEGL RiPROCESS, strip-adjusted, and exported as LAZ 1.4, including amplitude, reflectance, and pulse-shape deviation \citep{weiser2022syssifoss-essd}.

\begin{figure}[htbp]
  \centering
  \includegraphics[width=0.5\linewidth]{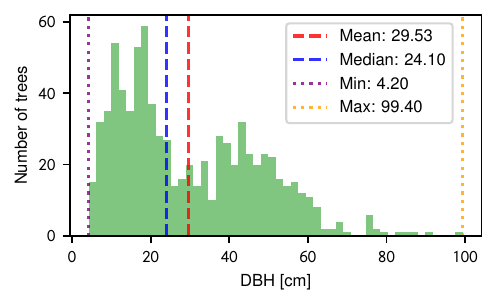}
  \caption{Distribution of diameter at breast height (DBH) values for all reference trees across all study sites. Vertical dashed lines mark the mean, median, minimum, and maximum DBH as indicated in the legend.}
  \label{fig:diameter-distribution}
\end{figure}

Field inventory measurements served as the basis for reference DBH values. DBH was measured with a diameter tape at 1.3~m above ground following standard practice. DBH was collected in close temporal proximity to align with ULS acquisitions. The DBH distribution across all employed plots is shown in Figure~\ref{fig:diameter-distribution}. In total, DBH values for 804 reference trees were collected across 20 species, comprising 14 broadleaf species (501 trees, dominated by \textit{Fagus sylvatica} and \textit{Quercus petraea}) and 6 conifer species (303 trees, dominated by \textit{Picea abies} and \textit{Pseudotsuga menziesii}).

\subsection{Individual tree segmentation}
\label{section:individual-tree-segmentation}

We employ an instance segmentation approach to isolate individual trees prior to DBH estimation. We use SegmentAnyTree \citep{wielgosz2024segmentanytree} without any dataset specific fine-tuning to evaluate applicability of an automatic pipeline and its transferability. SegmentAnyTree is a sparse tensor variation of the encoder-decoder style U-Net architecture widely used for semantic segmentation in computer vision. It employs parallel heads for semantic and instance segmentation, followed by region-growing and mean-shift clustering and a learned ScoreNet for candidate filtering and non-maximum suppression \citep{jiang2020pointgroup}.

To align predicted tree segments with ground truth instances for subsequent evaluation, we compute 3D axis-aligned bounding box Intersection over Union (IoU) between all ground truth and predicted instances within each plot and determine the optimal one-to-one assignment of predicted to reference trees using the Hungarian algorithm \citep{kuhn1955hungarian}, discarding only assignments with essentially no spatial overlap (IoU $<0.01$). For each matched tree, we additionally concatenate ground points within a 2.0~m radius around the trunk centroid to ensure adequate ground representation and reduce the effect of ground misclassification. The trunk centroid is derived as the center of points below 2.0~m height, under the assumption that these predominantly correspond to trunk points.

\begin{figure}[htbp]
  \centering
  \includegraphics{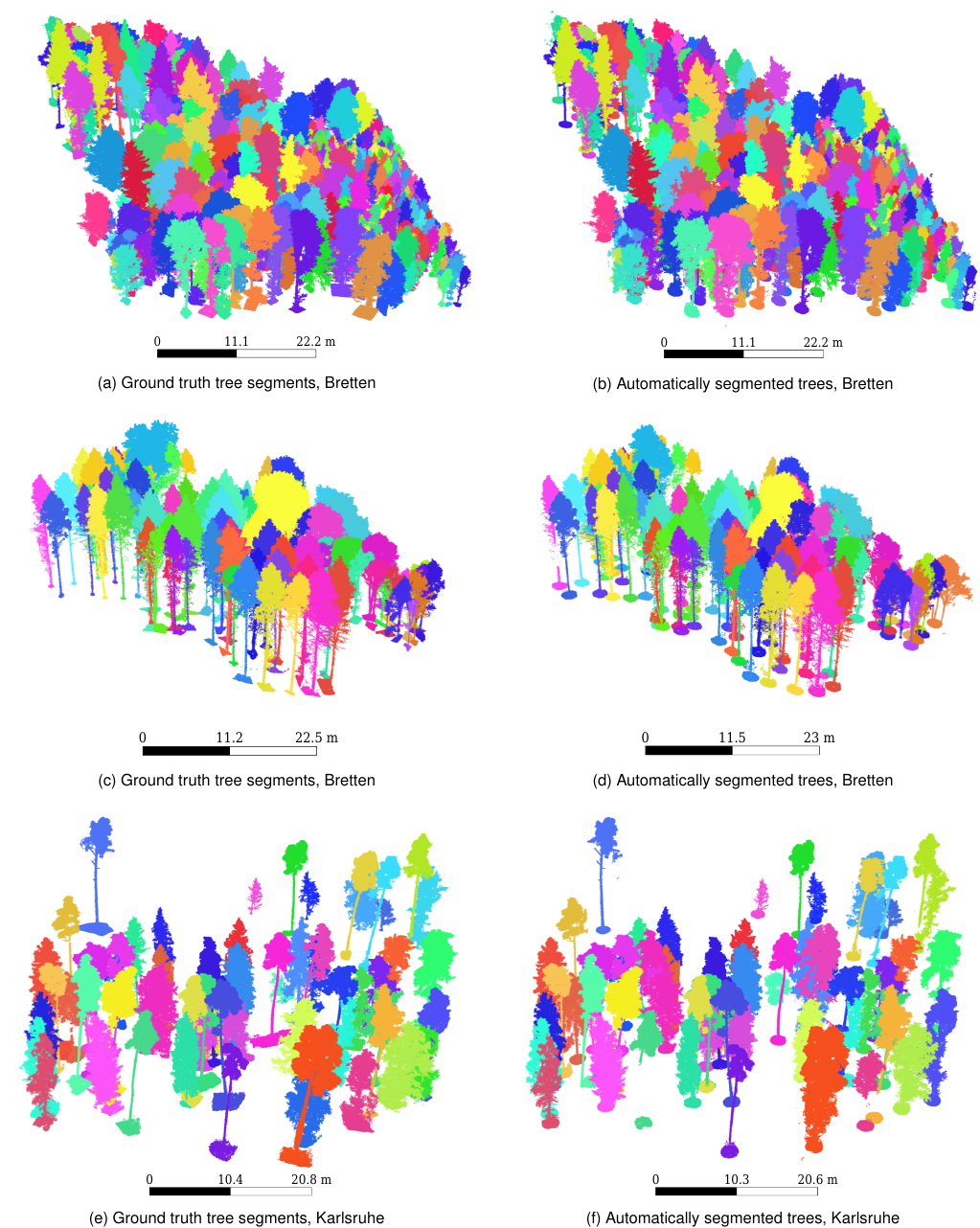}
  \caption{Visual comparison of ground truth and automatically segmented tree instances. (a,c,e) Ground truth tree instances. (b,d,f) Automatic instances from SegmentAnyTree using default inference settings. Colors indicate distinct tree instances.}
  \label{fig:tree-segments}
\end{figure}

Because automatic individual tree instance segmentation methods can incur missed detections, over- and under-segmentation, and inclusion of neighboring points, especially in dense, multilayered stands \citep{du2023lidar,yu2011predicting,martin2021evaluation}, many studies mitigate this by manually excluding smaller trees or difficult-to-segment instances \citep[e.g.][]{sun2022predicting,hao2020individual}. In contrast, we do not perform any manual filtering. To better evaluate the effect of errors that could potentially propagate from incorrect segmentation and bias the model ranking, we additionally evaluate our DBH models on expert-annotated ground truth single-tree instances. This allows us to quantify the upper bound attainable with perfect segmentation and isolate the impact of segmentation errors on DBH estimation. Visual examples of both automatically segmented trees and manual segments are shown in Figure~\ref{fig:tree-segments}.

\subsection{Tabular regression baselines}
\label{section:baselines}

A best practice for tree parameter regression from ULS established in recent studies consists of a two step approach, specifically extracting a comprehensive set of per-tree point cloud metrics (e.g., aggregated point height statistics and distributions) and fitting supervised tabular regressors to predict a tree parameter (e.g., DBH, above-ground biomass) \citep{hartley2025uav,xu2025improving,ye2025estimating,sumnall2023estimation,du2023lidar,sun2022predicting,zhang2022estimation}. This effectively frames the task as a typical tabular regression machine learning task, allowing the use of any strong supervised regressor, with typical choices being Linear Regression, Random Forest, and GBDTs like XGBoost. Because of this, we evaluate this approach as our primary baseline.

Linear Regression is a statistical method used to model the relationship between a dependent variable and one or more independent variables with a linear equation. Linear Regression is a fundamental baseline in tree parameter estimation in forestry applications \citep{xu2025improving,sumnall2023estimation,sun2022predicting}.

Random Forest \citep{breiman2001random} is a supervised ensemble learning method that constructs a large number of Decision Trees (DTs) and aggregates their individual predictions. Random Forest is a robust and high performing baseline, making it a popular choice for predicting forest attributes from LiDAR-derived point clouds \citep{hartley2025uav,xu2025improving,ye2025estimating}.

XGBoost \citep{chen2016xgboost} is a popular implementation of GBDT, an ensemble learning technique that builds a strong model by sequentially adding weak DTs, with each subsequent DT trained to correct prior errors. GBDTs, and XGBoost in particular, are historically among the best performing models in tabular machine learning applications, including forestry \citep{xu2025improving,ye2025estimating}.

\subsection{Sparse tensor networks}
\label{section:sparse-tensor-networks}

\begin{figure}[htbp]
  \centering
  \includegraphics[width=0.5\linewidth]{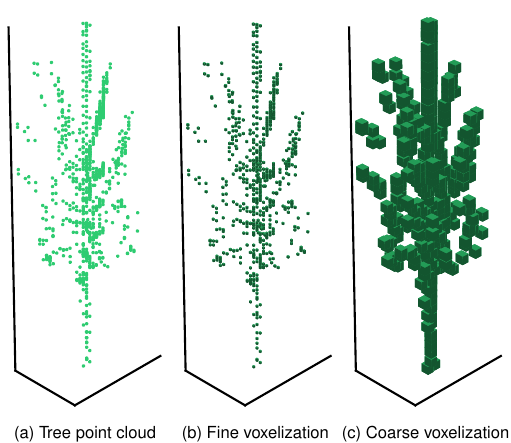}
  \caption{Illustration of voxelized tree representation. (a) Raw individual tree point cloud. (b) Fine voxelization (grid size 0.05 m). (c) Coarse voxelization (grid size 0.2 m).}
  \label{fig:voxelization}
\end{figure}

Sparse tensor networks, specifically sparse convolutional neural networks (CNNs), have emerged as a strong tool in LiDAR-based forestry research, and have recently achieved high accuracy for forest inventory tasks such as tree instance segmentation \citep{wielgosz2024segmentanytree, xiang2024automated}.

Sparse tensor networks operate on sparse tensors, a discretized representation of point cloud data that can be obtained through voxelization (see illustrative example in Figure~\ref{fig:voxelization}). This data representation allows storing only coordinates that are present in the point cloud, so the network only processes and produces values at those coordinates. This is enabled by generalized sparse convolutions, which accumulate features only from neighbors that actually exist, allow arbitrary kernel shapes and output coordinate sets, and reduce to ordinary convolutions if the coordinate sets cover the full grid \citep{graham2018submanifold,choy2019minkowskiengine}.

In contrast, standard convolutions used in regular CNNs (e.g., in image domains) assume dense grids and compute at every location with fixed kernels. For 3D LiDAR representations of forest point clouds, a dense representation would be prohibitively expensive in memory and compute due to volumetric scaling, while most of the space is empty. Exploiting sparsity is therefore critical. Sparse tensor CNNs make 3D convolutional processing feasible and well matched to the inherently sparse structure of LiDAR-derived forest point clouds.

Their strong performance for tree segmentation makes them a natural candidate to investigate for predicting individual tree parameters. However, applicability of sparse tensor CNNs, and even more generally deep learning and artificial neural networks, to single-tree parameter estimation and automated forest inventory has not yet been systematically explored \citep{rodrigues2024applications}. The primary study to directly explore point-cloud regression of forest attributes with sparse tensor networks is \citet{oehmcke2024deep}, who investigated sparse tensor networks for plot-level above-ground biomass and wood volume estimation on airborne laser scanning (ALS) subplots from the Danish National Forest Inventory. In their framework, sparse tensor variants of squeeze-and-excitation residual networks (MSENet14 and MSENet50) achieved the best results, outperforming approaches based on handcrafted point-cloud metrics. Our problem setting differs in two key aspects. First, we target single-tree rather than area-based parameter regression. Second, we operate on segmented individual trees from ULS, introducing a different spatial scale and point density regime than national-scale ALS subplots.

Nevertheless, the motivation and strong performance of sparse tensor networks introduced by \citet{oehmcke2024deep} are promising and applicable to our 3D forestry data. We therefore employ their proposed MSENet14 and MSENet50 models as strong sparse tensor network baselines in our study, adapting only task-specific details (e.g., voxel size and preprocessing). This allows us both to quantify the transferability of area-based sparse tensor CNNs to single-tree regression and to provide a fair point of comparison for our proposed architecture.

\subsection{Improved sparse tensor architecture}
\label{section:treenext-architecture}

\begin{figure*}[htbp]
  \centering
  \includegraphics[width=1\linewidth]{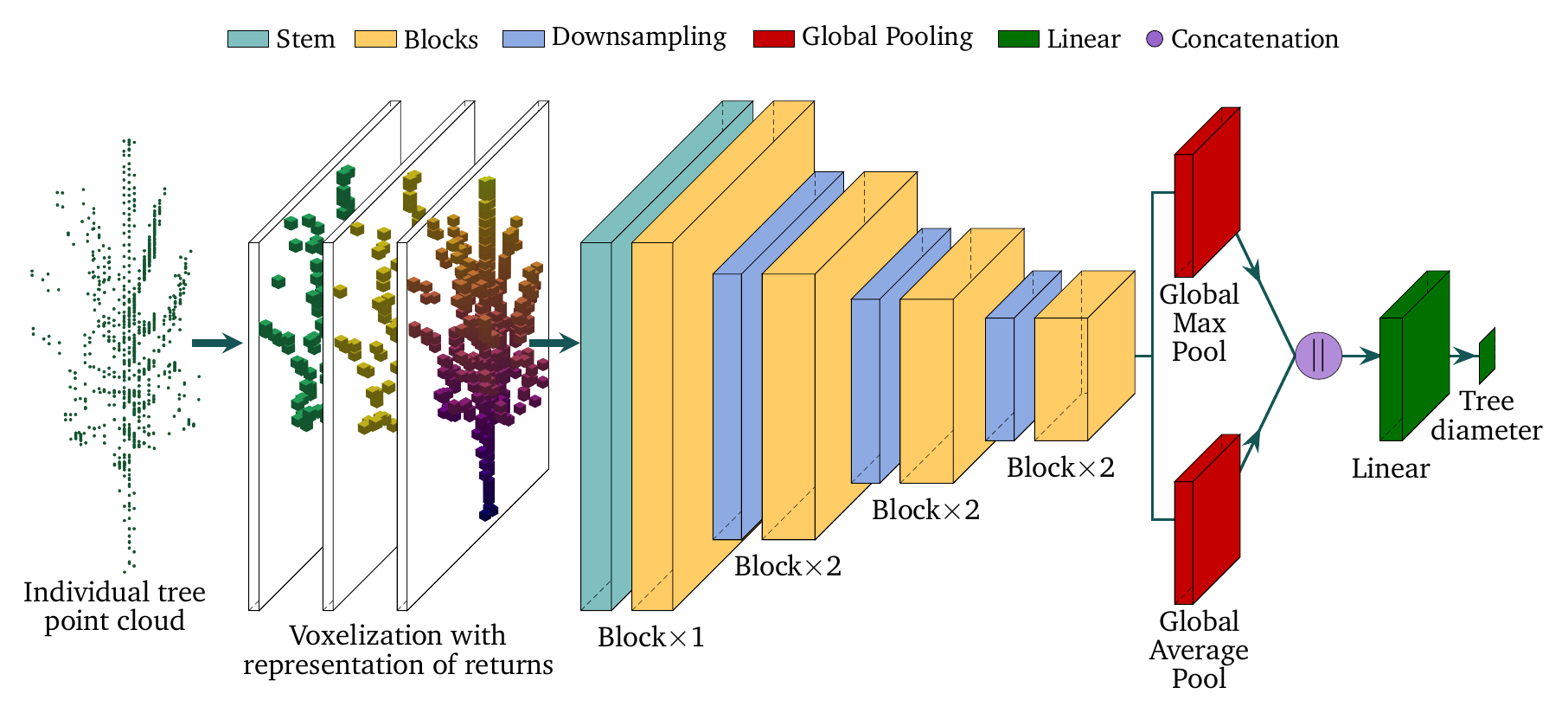}
  \caption{TreeNext network architecture. Individual tree point clouds are voxelized while preserving LiDAR return attributes as input channels. The model uses depthwise separable sparse convolutions with Global Response Normalization (GRN), hierarchical downsampling, and concatenated global max/average pooling.}
  \label{fig:treenext-architecture}
\end{figure*}

Building on the strengths of sparse tensor networks, we introduce TreeNext, a lightweight sparse tensor network architecture for single-tree parameter regression from LiDAR-derived point clouds. The architecture is shown in Figure~\ref{fig:treenext-architecture}. TreeNext introduces targeted improvements in data representation, network architecture, and training efficiency, aimed at improving both predictive performance and practicality for forestry applications.

LiDAR-derived point clouds typically carry pulse-level attributes, such as return number and number of returns. In forestry, this information can be particularly valuable, since the laser can penetrate vegetation and provide additional cues about 3D structure and material properties \citep{jamru2024exploring}. Because employing sparse tensor CNNs involves a voxelization step (in our case via grid sampling), prior work often discarded return-related attributes. We preserve this information by encoding it as additional input channels and aggregating the attributes within each voxel during voxelization.

We also introduce several architectural improvements that enhance both predictive performance and computational efficiency. TreeNext uses a sparse version of depthwise separable convolutions, originally introduced in MobileNet \citep{howard2017mobilenets}. We further improve the global pooling by using the concatenation of global max and global average pooling, which has been shown to capture complementary summary statistics compared to either pooling operation alone \citep{lee2016pooling}. Another particularly consequential design choice concerns normalization. Batch Normalization, used in prior work, fails to maintain stable batch statistics when memory constraints restrict the training batch size \citep{wu2018group}. We therefore replace Batch Normalization with Global Response Normalization (GRN) as introduced in ConvNeXt-V2 \citep{woo2023convnextv2}, which is batch-size agnostic and stabilizes training across different sparsity patterns. Beyond improving training stability, this choice enables the use of gradient accumulation \citep{goyal2017accurate}, a standard technique that reduces GPU memory consumption during training. This substantially lowers hardware requirements, operationally allowing the processing of higher point densities or finer voxel grid sizes on the same hardware. These design choices collectively yield a single lightweight architecture that is more memory efficient and faster to train and evaluate than the MSENet variants proposed by \citet{oehmcke2024deep}.

\subsection{Experimental setup and evaluation}
\label{section:evaluation}

For performance evaluation, we use the coefficient of determination ($R^2$):
$$
R^2 = 1 - \frac{\sum_{i=1}^{n}(y_i - \hat{y}_i)^2}{\sum_{i=1}^{n}(y_i - \bar{y})^2}
$$
where $y_i$ are the ground truth DBH values, $\hat{y}_i$ are the predictions, and $\bar{y}$ is the mean of the true values.

We additionally compute the root mean squared error (RMSE), used both for hyperparameter optimization and as a supplementary performance metric:
$$
\text{RMSE} = \sqrt{\frac{1}{n}\sum_{i=1}^{n}(y_i - \hat{y}_i)^2}
$$
We report $R^2$ as the primary metric in the main results and include RMSE in the extended results (Appendix~\ref{appendix:extended-results}). We additionally report the relative RMSE (rRMSE$\,= 100 \cdot \text{RMSE}\,/\,\bar{y}$), expressing RMSE as a percentage of the mean reference value.

For tabular regression baselines, we employ nested cross-validation with 5 outer folds and 5 inner folds for hyperparameter tuning (see Appendix~\ref{appendix:hyperparameter-grids}). Since these baselines require calculated point cloud metrics, we use a comprehensive set based on previous literature, outlined in Appendix~\ref{appendix:point-cloud-metrics} Table~\ref{tab:baseline_metrics}.

We train the sparse tensor networks using the same 5 outer folds as the tabular baselines to enable fair comparison. Within each outer training fold, an additional 10\% subset is held out from the training data to independently monitor model performance during training. Similar to the baselines, the outer test folds are fully independent and are only used for final performance evaluation. Training is done for 300 epochs with a batch size of 32, optimizing an L1 loss using the AdamW optimizer with a cosine annealing learning rate scheduler. Preprocessing employs voxel grid sampling with a fixed grid size of 0.05~m for all sparse tensor networks. Data augmentation is applied during training, including random z-axis rotations up to $360^\circ$ and random voxel dropout performed after grid sampling.

To account for variability during neural network training \citep{barkov2025benchmark}, we repeat the training and evaluation 5 times. We report mean and standard deviation of $R^2$ across these runs. We also evaluate a post-hoc ensemble of these repeated runs by averaging their predictions to showcase another practical advantage of employing artificial neural networks. For a fair comparison, we apply the same post-hoc ensembling strategy to the tabular baselines.

For TreeNext specifically, we conduct two additional analyses. First, to assess data efficiency, we evaluate model performance after reducing the available training data within each outer fold to fractions of 10\%, 20\%, 40\%, 60\%, while keeping the outer test folds unchanged. Second, we examine the effect of ensemble size by comparing $R^2$ improvements relative to a single model across ensemble sizes from $k = 2$ to $k = 16$.

\section{Results}
\label{section:results}

\begin{figure}[htbp]
  \centering
  \includegraphics{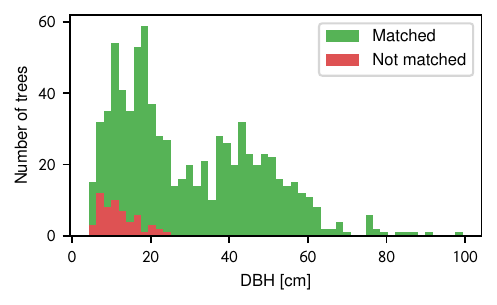}
  \caption{Distribution of reference tree DBH values by matching status after segmentation-based instance assignment. Matched trees (green) were successfully linked to a predicted segmented instance. Unmatched trees (red) had no corresponding instance. All unmatched trees have a DBH $<$ 25~cm.}
  \label{fig:tree-matching}
\end{figure}

Before evaluating DBH estimation performance, we quantify the outcome of the segmentation-based tree matching that links predicted instances to reference trees (Section~\ref{section:individual-tree-segmentation}). Of the 804 valid reference trees across all study plots, 747 (92.9\%) were successfully matched to a segmented instance, while 57 (7.1\%) remained unmatched. The distribution of matched and unmatched trees across the DBH range is shown in Figure~\ref{fig:tree-matching}.

The unmatched trees are exclusively small understory trees. All 57 unmatched trees have a DBH below 25~cm, with a mean of 11.82~cm (SD = 4.82~cm) and a maximum of only 23.70~cm. The majority of unmatched trees (75.4\%) have a DBH $<$ 15~cm. This is consistent with the known difficulty of detecting and segmenting small, suppressed trees in dense, multilayered stands from above-canopy ULS acquisitions. The match rate increases sharply with tree size: all trees with DBH $\geq25$~cm were matched at a rate of 100\%, and the rate remains above 97\% for trees with DBH $\geq15$~cm (97.7\%) and above 98\% for trees with DBH $\geq20$~cm (98.7\%).

Notably, even among small trees the automatic segmentation recovers the majority of them. Of the 82 reference trees with DBH below 10~cm, 59 (72.0\%) were successfully matched, and of the 201 trees below 15~cm, 158 (78.6\%) were recovered. These results indicate that the fully automatic segmentation pipeline captures the vast majority of the tree population, with missed detections concentrated exclusively among the smallest understory trees.

\begin{table*}[htbp]
  \centering
  \caption{Model performance comparison. The results are calculated after 5-fold cross validation. Mean and ensemble $R^2$ are calculated over repeated cross validation runs (k=5). Model parameter counts for sparse tensor networks are reported in millions (M).}
  \label{tab:model_performance}
  \begin{tabular}{l cc cc c}
    \toprule
    & \multicolumn{2}{c}{Automatically segmented trees}
    & \multicolumn{2}{c}{Ground truth segments} & Parameters \\
    \cmidrule(lr){2-3} \cmidrule(lr){4-5}
    Model & Mean $R^2$ & Ensemble $R^2$ & Mean $R^2$ & Ensemble $R^2$ & ~\\
    \midrule
    \multicolumn{6}{c}{\textit{Sparse Tensor Networks}} \\
    \addlinespace[2pt]
    TreeNext \textit{(Proposed)}
    & $\mathbf{0.8266\pm0.0018}$ & $\mathbf{0.8355}$
    & $\mathbf{0.9322\pm0.0027}$ & $\mathbf{0.9380}$ & 5.19~M \\
    MSENet50
    & $0.7872\pm0.0069$ & $0.8004$
    & $0.9076\pm0.0045$ & $0.9157$ & 48.71~M \\
    MSENet14
    & $0.7893\pm0.0025$ & $0.7963$
    & $0.8944\pm0.0028$ & $0.8996$ & 14.40~M \\
    \addlinespace[4pt]
    \multicolumn{6}{c}{\textit{Tabular Regression with Point Cloud Metrics}} \\
    \addlinespace[2pt]
    XGBoost
    & $0.8045\pm0.0031$ & $0.8086$
    & $0.9296\pm0.0006$ & $0.9315$ & --- \\
    Random Forest
    & $0.8024\pm0.0009$ & $0.8027$
    & $0.9219\pm0.0003$ & $0.9220$ & --- \\
    Linear Regression
    & $0.7905$ & ---
    & $0.9283$ & --- & --- \\
    \bottomrule
  \end{tabular}
\end{table*}

\begin{figure*}[htbp]
  \centering
  \includegraphics[width=1.0\linewidth]{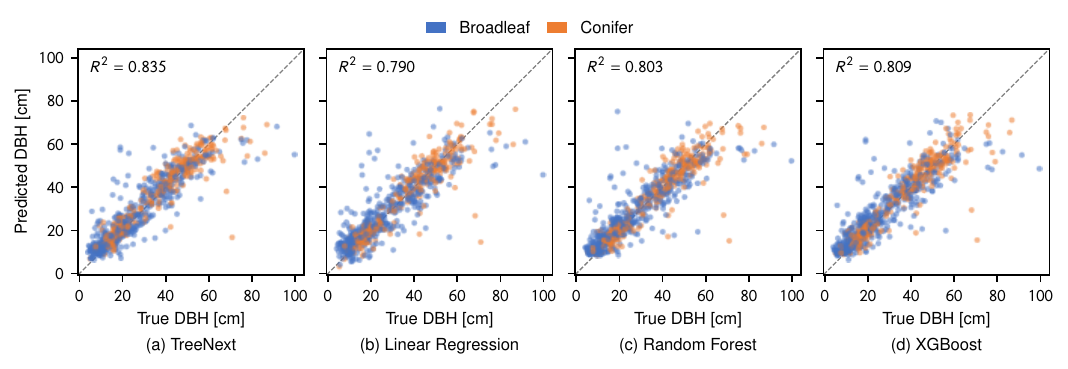}
  \caption{Scatter plots of predicted versus reference DBH for TreeNext and the tabular regression baselines on automatically segmented trees. Points are colored by broadleaf and conifer species group. The dashed line indicates the 1:1 relationship.}
  \label{fig:scatter-plots}
\end{figure*}

After examining the segmentation-based tree matching, we evaluate DBH estimation performance. Table~\ref{tab:model_performance} summarizes the cross-validated performance, with corresponding scatter plots of predicted versus reference DBH provided in Figure~\ref{fig:scatter-plots}. Across both segmentation settings, TreeNext is the best performing model with low variance across runs. Compared to the strongest tabular machine learning baseline, XGBoost, TreeNext improves average $R^2$ from $0.8045$ to $0.8266$ on automatically segmented trees and from $0.9296$ to $0.9322$ on ground truth segments. Post-hoc ensembling provides additional improvement for TreeNext, whereas tree-based models see limited benefit. Notably, the sparse tensor baselines adapted from previous work on plot-level tasks, MSENet14 and MSENet50, show reduced performance compared to all tabular machine learning baselines in both evaluation settings, while TreeNext performs well in all scenarios. Overall, TreeNext demonstrates robustness and high performance across all conditions.

Switching from evaluating on automatically segmented trees to ground truth tree segments yields substantial increases for all models, indicating that segmentation errors are a dominant source of performance loss in the end-to-end automated tree inventory pipeline. TreeNext preserves its ranking in both scenarios, with a notably larger relative margin when segmentation is automatic.

Figure~\ref{fig:scatter-plots} shows predicted versus reference DBH for TreeNext and the tabular regression baselines on automatically segmented trees. Compared to TreeNext, tabular baselines exhibit higher scatter across the full DBH range, particularly for trees with DBH above approximately 50~cm. Broadleaf and conifer trees span the full diameter range across all models, with scatter generally slightly increasing at larger diameters regardless of species group. However, some systematic bias between species groups that is consistent across all four models can also be noticed: the largest individual overestimation cases, specifically points above the 1:1 line, tend to be predominantly broadleaf trees.

\begin{figure}[htbp]
  \centering
  \includegraphics[width=0.5\linewidth]{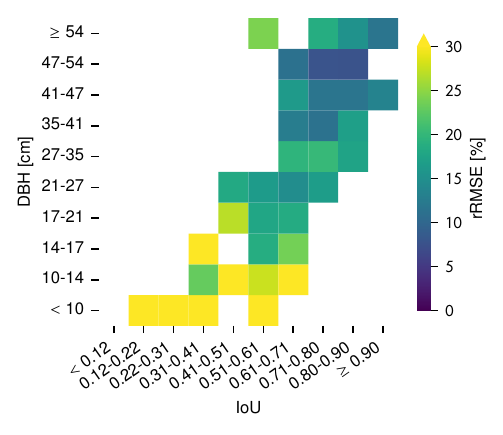}
  \caption{Heatmap of TreeNext DBH estimation error (rRMSE, \%) as a function of segmentation IoU (across 10 uniform bins) and ground truth DBH (across 10 quantile bins). Bins with at least ten trees are shown.}
  \label{fig:iou-vs-dbh-vs-rmse}
\end{figure}

Figure~\ref{fig:iou-vs-dbh-vs-rmse} illustrates the relationship between segmentation quality (IoU), ground truth tree diameter (DBH), and the resulting DBH estimation relative error (rRMSE). Small trees (DBH below 10--17~cm) suffer from the highest relative errors, reaching rRMSE values above 25\%. Crucially, these high relative errors coincide with the lowest segmentation quality: small trees show lowest IoU values, being simultaneously the hardest to segment and the hardest to estimate accurately relative to their size. Conversely, large, well-segmented trees achieve the lowest rRMSE values, demonstrating that both good segmentation and larger tree size contribute to accurate relative estimation. In general, regardless of the tree size, rRMSE consistently decreases as IoU increases across all DBH bins and remains comparatively low once IoU exceeds about 0.5.
\begin{figure*}[htbp]
  \centering
  \includegraphics[width=1.0\linewidth]{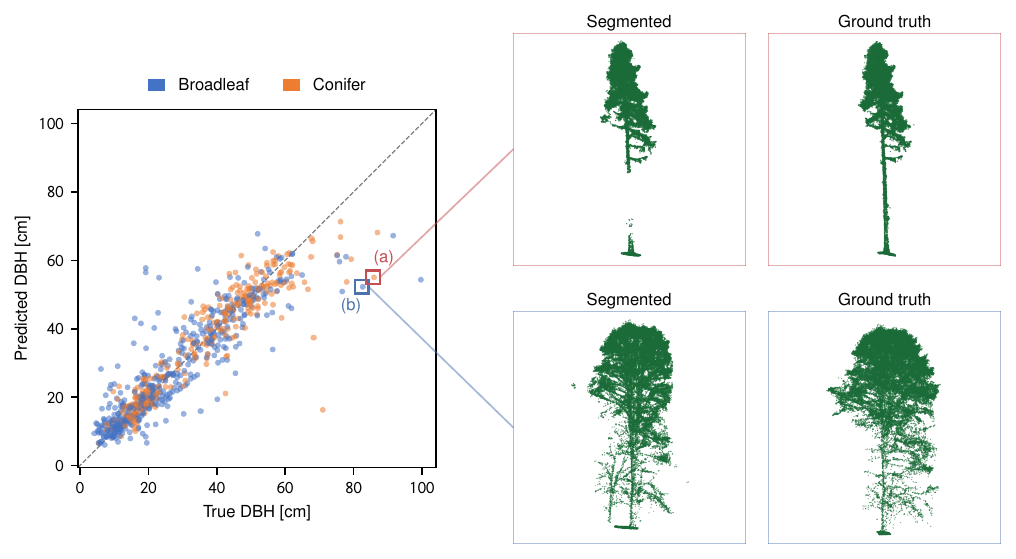}
  \caption{Illustration of automatic segmentation issues of two selected trees with high DBH estimation error. The scatter plot shows predicted versus reference DBH for TreeNext on automatically segmented trees. Points are colored by broadleaf and conifer species group. Two outliers are highlighted: (a) a \emph{Larix decidua} conifer tree, and (b) a \emph{Fagus sylvatica} broadleaf tree. For each outlier, the automatically segmented point cloud and the corresponding ground truth segment are shown.}
  \label{fig:outlier-inspection}
\end{figure*}

Figure~\ref{fig:outlier-inspection} highlights two representative cases where segmentation artifacts are visible for specific trees with large DBH estimation errors. In the \emph{Larix decidua} example (a), automated segmentation produced a fragmented instance that retains only the upper crown while the stem region is largely absent from the segment. In the \emph{Fagus sylvatica} example (b), the automated segment incorporates neighboring understory vegetation. In both cases, the large discrepancy between the automatic segment and the corresponding ground truth coincides with substantial DBH estimation error, consistent with the broader pattern shown in Figure~\ref{fig:iou-vs-dbh-vs-rmse}.

\begin{figure}[htbp]
  \centering
  \includegraphics[width=0.5\linewidth]{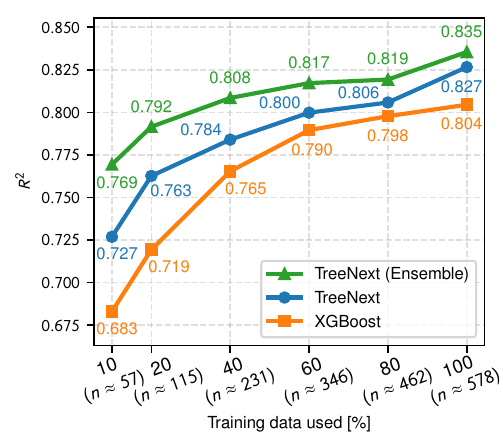}
  \caption{Effect of training set size on performance for TreeNext, its post-hoc ensemble (TreeNext Ensemble), and the strongest tabular baseline (XGBoost) across fractions of the available training data. Sample counts correspond to one outer training fold in cross validation.}
  \label{fig:ablate-dataset-size}
\end{figure}

We further investigated the sensitivity of the proposed model to training set size, as shown in Figure~\ref{fig:ablate-dataset-size}. TreeNext already outperforms the strongest tabular baseline (XGBoost) at every training data fraction, and the post-hoc ensemble strengthens this advantage further. In the low-data regime the advantage is pronounced: at $10\%$ of the data ($\approx57$ samples) TreeNext reaches $R^2=0.727$ vs.\ $0.683$ for XGBoost ($+0.044$), while the ensemble reaches $R^2=0.769$ at the same fraction ($+0.086$ over XGBoost). The ensemble configuration already exceeds the full-data XGBoost baseline ($R^2=0.804$) with only $40\%$ of the training data ($R^2=0.808$). Performance gains for the individual TreeNext model diminish beyond $60\%$ of the data, while the ensemble continues to benefit from additional training samples, reaching $R^2=0.835$ at full data. This demonstrates that TreeNext is highly data efficient, and that the ensemble configuration amplifies this advantage substantially.

Finally, we evaluated the effect of post-hoc ensembling across repeated TreeNext runs, as shown in Appendix~\ref{appendix:ablate-ensemble}. Predominantly, post-hoc ensembling provides consistent additional performance improvements and positions TreeNext even stronger against all baselines. For automatically segmented trees, ensembling $k=5$ TreeNext models improves $R^2$ by $+0.0089$ (from $0.8266$ to $0.8355$). However, gains saturate quickly: the marginal benefit peaks at $k=2$ ($\approx+0.008$ $R^2$) and then decays, with increases beyond $k\approx5$ being really small ($<0.002$ absolute).

Overall, TreeNext delivers the best DBH regression from ULS point clouds across both segmentation settings, with the advantage over the strongest tabular baseline (XGBoost) being particularly pronounced on automatically segmented trees. Better instance segmentation would strongly improve all methods, with TreeNext remaining the best performer. The model is highly data-efficient, outperforming XGBoost at all training data fractions, and the post-hoc ensemble further amplifies this advantage.

\section{Discussion}
\label{section:discussion}

Properly informing operational forest management and ecological research with detailed single-tree information in an era of rapidly expanding remote sensing capabilities calls for robust, performant models that can extract maximum information from LiDAR point clouds and fully leverage costly and labor-intensive field campaigns. Our investigation demonstrates that existing approaches to estimate DBH from ULS data can be improved by applying  deep learning workflows, with sparse tensor networks bringing improved performance to automated forest inventory from ULS point clouds. The operational flexibility of ULS, when coupled with the lightweight design and strong predictive accuracy of TreeNext, enables efficient estimation of tree parameters such as DBH at individual tree level, directly supporting allometric modeling of biomass, carbon, and timber volume and, thereby, strengthening the foundation of forest management planning in precision forestry.

The most direct advantage of applying sparse tensor networks to forest inventory is improved predictive performance. In our evaluation, TreeNext in the ensemble configuration achieved an $R^2$ of 0.8355 on automatically segmented trees and 0.9380 on ground truth segments, compared to the strongest tabular baseline (XGBoost ensemble) of 0.8086 on automatically segmented trees and 0.9315 on ground truth segments. While these improvements are meaningful for forestry applications, the significance of this result extends beyond individual point-estimate accuracy. Learning directly from voxelized 3D point clouds rather than predefined handcrafted metrics represents a qualitative shift in how forest inventory models can be developed and refined further.

Beyond better predictions, one of the notable findings of our study is the strong performance of TreeNext even with limited availability of ground truth data. The application of machine learning has brought substantial gains to scaling forest inventory, yet it encounters the same obstacle as any other environmental sensing application that relies on supervised learning: the need for labor-intensive, time-consuming, and costly labeled reference data \citep[e.g.][]{lau2019estimating,moser2017methods,barkov2025benchmark,schmidinger2025limesoda}. Deep learning in particular, while driving many recent advances, is commonly perceived as data-hungry \citep{esmoris2024deep}, with performance scaling primarily with dataset size \citep{wang2023tree}. Our results challenge this assumption: TreeNext outperforms XGBoost, the strongest tabular baseline, at every training data fraction, and the ensemble configuration already exceeds the full-data XGBoost baseline with only 40\% of the training data. While recent work has shown that tabular models can be calibrated with as few as 10--55 tree samples \citep{brede2022nondestructive}, our results demonstrate that a deep learning approach can extract even more information from the same limited data, achieving substantially higher accuracy (e.g., $R^2 = 0.727$ for a single model and $R^2 = 0.769$ with the ensemble, compared to $0.683$ for XGBoost, with approximately 57 samples). This data efficiency is a strong practical advantage for operational forestry applications and removes one of the perceived barriers to adopting deep learning methods in routine forest inventory.

Beyond data efficiency, moving toward actual operational deployment and taking advantage of ULS flexibility for fast, repeated acquisition campaigns that support forest ecosystem monitoring and carbon sequestration estimation \citep{shao2024largescale} requires that the computational demand of newly introduced approaches remains tractable. Here our approach is equally advantageous. In practice, we observed that MSENet50, the larger sparse tensor baseline adopted from prior work \citep{oehmcke2024deep}, required approximately 57.9~GB of GPU memory during training and 7.6~GB of GPU memory during inference under our experimental settings, making it somewhat limiting for routine operational use. In contrast, the lightweight architecture of TreeNext (5.19~M parameters vs.\ 48.71~M for MSENet50), combined with architectural choices that enable modern memory-efficient training techniques such as gradient accumulation, can reduce GPU memory requirements to approximately 5.9~GB during training and 0.5~GB during inference. This is an advantage for operational forestry that extends beyond the current experiments: improved performance does not come at the cost of operational feasibility, establishing a direct path toward practical deployment.

Inventory from airborne platforms has been rapidly adopted by the forestry community, with area-based approaches being among the first to reach operational maturity \citep{coops2021modelling}. However, for management decisions that require fine-grained structural detail, such as selective thinning prescriptions, competition assessment, and tree-level growth projections, single-tree information is essential \citep{sun2022predicting,kukkonen2022evaluation,harrington2011stand}. Indeed, as forest inventories increasingly produce tree-level data, a logical aim is to employ these data directly in management planning \citep{sun2022predicting}, and individual tree biomass estimation can advance more accurate forest carbon accounting \citep{xu2021lidar}. TreeNext was evaluated precisely in this context, delivering strong DBH estimation at the individual tree scale. Critically, the entire workflow from segmentation to parameter estimation operates within a deep learning framework without manual intervention, with both stages relying on sparse tensor networks. As inventories scale toward larger areas and higher acquisition frequencies, manual correction or curation of intermediate results becomes increasingly impractical. The ability to demonstrate competitive results within a fully automated pipeline is therefore especially valuable, supporting the transition from small plot-based surveys to comprehensive, large-scale forest analysis \citep{goodbody2018assessing}.

Within this automated pipeline, tree segmentation emerges as both the current performance bottleneck and a particularly promising avenue for future improvement. During instance segmentation, 7.1\% of reference trees could not be matched to a predicted segment, with omissions concentrated among the smallest understory individuals: 75.4\% of unmatched trees had a DBH below 15~cm, and all fell below 25~cm. In context, this is a favorable outcome: \citet{du2023lidar} reported that approximately 35\% of delineated crowns could not be matched to field-measured trees, and \citet{hyyppa2022direct} found completeness rates of only 20--30\% for stems with DBH smaller than 20~cm. Moreover, since the omissions represent predominantly small, suppressed understory trees, they may have limited impact on many ecological and management applications where such individuals can reasonably be treated as part of a larger structural unit rather than discrete objects \citep{jeronimo2018applyinga}, and where competition indices and growth models typically focus on dominant and codominant stems \citep{barua2025predicting}. Nevertheless, the omissions are present and point toward a clear direction for improvement that should be pursued in future work.

Besides detecting tree instances, the quality of segmentation itself influences downstream estimation accuracy. The clear relationship between segmentation quality and DBH estimation error (Figure~\ref{fig:iou-vs-dbh-vs-rmse}), together with the substantial improvement observed when evaluating on ground truth segments (Table~\ref{tab:model_performance}), identifies segmentation fidelity as another important avenue for future gains. A straightforward option is dataset-specific fine-tuning of the segmentation network \citep[e.g.][]{cherlet2026benchmarking}: we employed SegmentAnyTree with pretrained weights and default settings to assess out-of-the-box transferability, but recent research on model transferability in forestry suggests that even limited local calibration can substantially close the gap toward ground truth segmentation quality \citep{soininen2025transferability}. Of course, in practice, tree segments will inevitably be affected by sensor noise, processing artifacts, varying flight conditions, and environmental factors. The fact that TreeNext maintains its lead over all baselines, and with a larger relative margin, on realistic, automatically obtained tree segments is a definite advantage for real-world operational deployment.

Several directions for future research emerge from these findings. An interesting avenue for operational deployment is incorporating predictive uncertainty into forestry decision-making, enabling practitioners to flag low-confidence estimates for follow-up field verification \citep[e.g.][]{barkov2024efficient,kristiadi2021laplace}. Moreover, the strong performance of TreeNext should not overshadow the continued value of tabular baselines. Random Forest and XGBoost remained competitive on ground truth segments and offer computational simplicity and transparent variable importance that make them accessible operational tools. These models may also benefit from recent advances in feature engineering, such as topological descriptors based on simplicial complexes \citep{vonwestenholz2025simplicial}, or from hybrid architectures that fuse handcrafted metrics with deep learning features \citep{murray2025individual}. The lightweight design of TreeNext additionally opens a direct path toward scaling to even denser point cloud sources such as MLS, where the number of points per tree can be substantially larger than in ULS acquisitions but the platform remains operationally viable for detailed stand-level surveys. Consistent with a large body of laser scanning research \citep{kokeza2026influence}, our evaluation was conducted under leaf-off conditions, which favorably reduce vegetation occlusion. Whether the performance of TreeNext transfers to leaf-on acquisitions remains a natural direction for future work. Furthermore, while our evaluation demonstrates strong results across mixed Central European temperate forests, validation across additional forest types (e.g., boreal, tropical), and management regimes, including the structurally uniform conditions of intensive forest plantations \citep{kanninen2010plantation,gavilan2025use}, remains an important step toward establishing broader generalizability and guiding wider adoption in operational forest inventory.

\section{Conclusion}
\label{section:conclusion}

Sparse tensor networks provide an effective approach for direct individual tree DBH estimation from ULS point clouds. Building on the prior work that established their strength for plot-level biomass estimation from ALS data, we introduced TreeNext, a lightweight sparse tensor network for forestry applications. TreeNext showed the highest performance for single-tree DBH estimation in Central European mixed temperate forests among all evaluated models, with an $R^2$ of 0.8355 on automatically segmented trees and $R^2$ of 0.9380 when evaluated directly on ground truth tree segments.

Segmentation quality emerged as the primary driver of error in the automated forest inventory pipeline. All evaluated models showed reduced performance on automatically segmented trees, with relative error consistently highest for smaller trees. Improving individual tree instance segmentation, especially for small understory trees in dense, multilayered stands, represents the most direct step toward achieving meaningful improvement in end-to-end automated single-tree inventory from ULS data.

While the successful application of deep learning often relies on the availability of large labeled datasets, our experiments show that sparse tensor networks can be highly data-efficient. TreeNext outperformed the strongest baseline, XGBoost, even when trained with limited data, and its post-hoc ensemble exceeded the performance of XGBoost trained on the full dataset even when using only 40\% of the available training data.

We believe that this combination of strong predictive performance, data efficiency, and low computational requirements makes sparse tensor networks, and TreeNext in particular, a practical foundation for automated tree parameter estimation from ULS point clouds, and a valuable tool in the toolbox of every forest practitioner.

\section*{Acknowledgements}
\label{section:acknowledgements}

The authors thank Prof. Dr. Tim Römer for reviewing the manuscript and providing valuable feedback.

\section*{Author contributions}
\label{section:author-contributions}

Clara L\"{o}\ss l (Conceptualization, Investigation, Data curation, Methodology, Software, Validation, Formal analysis, Visualization, Writing--original draft, Writing--review \& editing), %
Fabian Ewald Fassnacht (Investigation, Writing--review \& editing), and %
Viacheslav Barkov (Conceptualization, Investigation, Data curation, Methodology, Software, Validation, Formal analysis, Visualization, Writing--original draft, Writing--review \& editing)

\section*{Conflict of interest}
\label{section:conflicts-of-interest}

None declared.

\section*{Funding}
\label{section:funding}

This work was supported by the Lower Saxony Ministry of Science and Culture (MWK), via the zukunft.niedersachsen program of the Volkswagen Foundation (ZN4072). Compute resources were supported by the Deutsche Forschungsgemeinschaft (DFG, German Research Foundation) project number 456666331.

\section*{Data availability}
\label{section:code-and-data-availability}

Code that supports the findings of this study is openly available at \url{https://github.com/slavabarkov/treenext}. The data that support the findings of this study are openly available in PANGAEA Repository at \url{https://doi.org/10.1594/PANGAEA.942856}.

\clearpage

\appendix

\section{Effect of ensemble size}
\label{appendix:ablate-ensemble}

\begin{figure*}[htbp]
  \centering
  \includegraphics[width=1\linewidth]{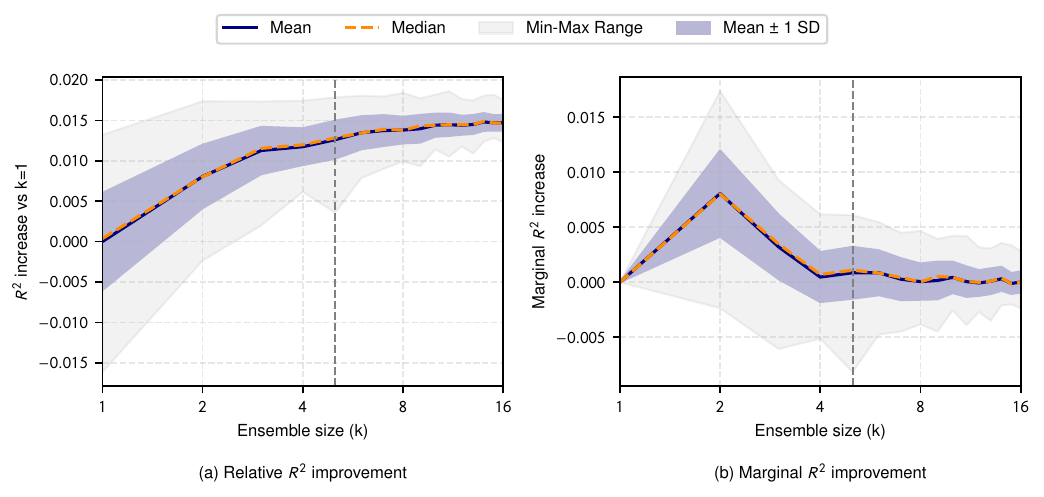}
  \caption{Effect of post-hoc ensemble size on TreeNext. (a) Relative improvement compared to a single model (k=1). (b) Marginal gain when increasing k. Shaded bands show $\pm1$~SD across repeated runs. The vertical dashed line marks k=5 used in the main results.}
  \label{fig:ablate-ensemble}
\end{figure*}

Figure~\ref{fig:ablate-ensemble} illustrates the effect of post-hoc ensemble size on TreeNext, reporting both the relative $R^2$ improvement over a single model and the marginal gain when increasing the ensemble size $k$.

\section{Extended results}
\label{appendix:extended-results}

\begin{table}[htbp]
  \centering
  \caption{Model performance comparison. The results are calculated after 5-fold cross validation. Mean RMSE and relative RMSE (rRMSE) are calculated over repeated cross validation runs (k=5).}
  \label{tab:results-extended-rmse}
  \begin{tabular}{l cc cc}
    \toprule
    & \multicolumn{2}{c}{Automatically segmented trees}
    & \multicolumn{2}{c}{Ground truth segments} \\
    \cmidrule(lr){2-3} \cmidrule(lr){4-5}
    Model & RMSE & rRMSE (\%) & RMSE & rRMSE (\%)\\
    \midrule
    \multicolumn{5}{c}{\textit{Sparse Tensor Networks}} \\
    \addlinespace[2pt]
    TreeNext \textit{(Proposed)}
    & $\mathbf{7.3246\pm0.0388}$ & $\mathbf{23.72\pm0.13}$
    & $\mathbf{4.6074\pm0.0915}$ & $\mathbf{15.60\pm0.31}$ \\
    MSENet50
    & $8.1143\pm0.1325$ & $26.27\pm0.43$
    & $5.3767\pm0.1308$ & $18.21\pm0.44$ \\
    MSENet14
    & $8.0751\pm0.0485$ & $26.15\pm0.16$
    & $5.7497\pm0.0769$ & $19.47\pm0.26$ \\
    \addlinespace[4pt]
    \multicolumn{5}{c}{\textit{Tabular Regression with Point Cloud Metrics}}\\
    \addlinespace[2pt]
    XGBoost
    & $7.7777\pm0.0628$ & $25.18\pm0.20$
    & $4.6937\pm0.0211$ & $15.89\pm0.07$ \\
    Random Forest
    & $7.8193\pm0.0173$ & $25.32\pm0.06$
    & $4.9438\pm0.0091$ & $16.74\pm0.03$ \\
    Linear Regression
    & $8.0520$ & $26.07$
    & $4.7372$ & $16.04$ \\
    \bottomrule
  \end{tabular}
\end{table}

Table~\ref{tab:results-extended-rmse} provides extended performance metrics for the models evaluated in the main study, specifically root mean squared error (RMSE) and relative RMSE (rRMSE).

\section{Tabular baseline point cloud metrics}
\label{appendix:point-cloud-metrics}

\begin{table}[htbp]
  \centering
  \caption{List of LiDAR point-cloud metrics extracted for each individual tree and used as input features for the tabular baseline models.}
  \label{tab:baseline_metrics}
  \begin{tabular}{{p{0.25\linewidth} p{0.05\linewidth} p{0.55\linewidth}}}
    \hline
    Category & \# & Definition \\
    \hline
    Basic height stats & 9 & Number of points, $z_{\min}$, $z_{\max}$, mean, variance, SD, $z_{cv}=100\cdot \text{SD}/\text{mean}$, skewness, and kurtosis computed from all return heights. \\
    \hline
    Percentiles & 21 & Height percentiles: 1st; 5th--95th (step 5); 99th. \\
    \hline
    Canopy density (cumulative) & 9 & Cumulative percentage of returns in 10 equal-width height bins covering the range $[z_{\min}, z_{\max}]$. \\
    \hline
    Dispersion & 5 & IQR; mean and median absolute deviations; canopy relief ratio $(\text{mean} - z_{\min})/(z_{\max} - z_{\min})$; Shannon entropy of $z>0$ using 1~m vertical layers. \\
    \hline
    Echo-based (aggregate) & 13 & Counts and percentages of first, intermediate, last, single, and multiple returns; ratios of last-to-first, intermediate-to-first, and multiple-to-single returns. \\
    \hline
    Echo-based (per return order) & var. & Point count $n_r$ for each return order $r$ present in the cloud ($r = 1,2,\dots$). \\
    \hline
    Exceedance & 3 & Percentage of returns above the mean height, and above fixed thresholds of 2~m and 5~m. \\
    \hline
    Height intervals & 8 & Percentage of returns within predefined absolute height bins (m). \\
    \hline
    Leaf Area Density (LAD) & 5 & Summary of LAD profile derived via Beer-Lambert law ($dz=1$~m, $k=0.5$, $z_0=2$~m): min, max, mean, $CV=\text{SD}/\text{mean}$, and total sum. \\
    \hline
    L-moments & 7 & First four L-moments and corresponding L-ratios. \\
    \hline
  \end{tabular}
\end{table}

Table~\ref{tab:baseline_metrics} details the employed point cloud metrics extracted from individual tree LiDAR point clouds and used as input features for the tabular regression baselines (Linear Regression, Random Forest, and XGBoost).

\section{Hyperparameter search space of tabular baselines}
\label{appendix:hyperparameter-grids}

\begin{table}[htbp]
  \centering
  \caption{Hyperparameter grids for classical baselines. Bracketed intervals were sampled uniformly; "log" indicates log-uniform sampling.}
  \label{table:classical-ml-hyperparameters}
  \begin{tabular}{lll}
    \toprule
    \textbf{Algorithm} & \textbf{Parameter} & \textbf{Search grid} \\
    \midrule
    XGBoost & learning\_rate & $[0.001,\,0.3]_{\log}$ \\
    & max\_depth & $\{3,\dots,10\}$ \\
    & min\_child\_weight & $\{1,\dots,10\}$ \\
    & subsample & $[0.5,\,1.0]$ \\
    & colsample\_bytree & $[0.5,\,1.0]$ \\
    & gamma & $[0,\,5]$ \\
    & reg\_alpha & $[10^{-6},\,10]_{\log}$ \\
    & reg\_lambda & $[0.1,\,10]_{\log}$ \\
    \midrule
    Random\,Forest & max\_depth & $\{3,\dots,30\}$ \\
    & min\_samples\_split & $\{2,\dots,10\}$ \\
    & min\_samples\_leaf & $\{1,\dots,10\}$ \\
    & max\_features & $[0.6,\,1.0]$ \\
    \bottomrule
  \end{tabular}
\end{table}

Table~\ref{table:classical-ml-hyperparameters} defines the hyperparameter search spaces employed for the Random Forest and XGBoost baselines. Optimal hyperparameters were selected via nested 5-fold cross-validation (random search with 100 iterations) on the outer training folds.

\clearpage

\bibliographystyle{Foresj}
\bibliography{references}

\end{document}